%% file: 0_main.tex
\documentclass[conference]{IEEEtran}
\IEEEoverridecommandlockouts

\usepackage{amsmath,amssymb,amsfonts}
\usepackage{graphicx}
\usepackage{adjustbox}
\usepackage{import}
\usepackage{cite}
\usepackage{wrapfig}
\usepackage{algorithm}
\usepackage{algpseudocode}
\usepackage{textcomp}
\usepackage{xcolor}
\usepackage{svg}
\usepackage{subfig}
\usepackage{soul}
\usepackage{multirow}
\usepackage{booktabs}
\usepackage{enumitem}
\usepackage{hyperref}
\usepackage{stfloats} 
\setlist[itemize]{noitemsep, topsep=0pt}

\long\def\/*#1*/{}

\newcommand\length{\texttt{L}}
\newcommand\width{\texttt{W}}
\newcommand{\MSE}{\texttt{MSE}}

\newcommand\asp{\texttt{asp}}

\def\generator{G}
\def\discriminator{D}

\def\BibTeX{{\rm B\kern-.05em{\sc i\kern-.025em b}\kern-.08em
    T\kern-.1667em\lower.7ex\hbox{E}\kern-.125emX}}

\title{Conditional Generative Models for High-Resolution Range Profiles: Capturing Geometry-Driven Trends in a Large-Scale Maritime Dataset
\\
\thanks{Submitted at Eusipco26. This work was performed using HPC resources from GENCI-IDRIS (Grant 2024-AD011014422R1).}}
\author{\IEEEauthorblockN{Edwyn Brient\IEEEauthorrefmark{1}\IEEEauthorrefmark{2}}
\IEEEauthorblockA{\textit{STIM, Mines Paris, PSL University}\\
Fontainebleau, France\\ edwyn.brient@minesparis.psl.eu}
\and
\IEEEauthorblockN{Santiago Velasco-Forero\IEEEauthorrefmark{1}}
\IEEEauthorblockA{\textit{STIM, Mines Paris, PSL University}\\
Fontainebleau, France\\ santiago.velasco@minesparis.psl.eu}
\and
\IEEEauthorblockN{Rami Kassab\IEEEauthorrefmark{2}}
\IEEEauthorblockA{\textit{ARC, Thales Land and Air Systems}\\
Limours, France\\ rami.kassab@thalesgroup.com}}

\begin{document}
%
\maketitle
\input{1_abstract}

\input{2_introduction}
\input{3_related}

\input{4_method}

\input{5_results}

\input{6_conclusion}

\bibliographystyle{IEEEbib}
\bibliography{biblio}

\end{document}

%% file: 1_abstract.tex
\begin{abstract} 
High-resolution range profiles (HRRPs) enable fast onboard processing for radar automatic target
recognition, but their strong sensitivity to acquisition conditions limits robustness across
operational scenarios. Conditional HRRP generation can mitigate this issue, yet prior studies are
constrained by small, highly specific datasets. We study HRRP synthesis on a large-scale maritime
database representative of coastal surveillance variability. Our analysis indicates that the
fundamental scenario drivers are geometric: ship dimensions and the desired aspect angle.
Conditioning on these variables, we train generative models and show that the synthesized
signatures reproduce the expected line-of-sight geometric trend observed in real data. These
results highlight the central role of acquisition geometry for robust HRRP generation.
\end{abstract}



%% file: 2_introduction.tex
\section{Introduction}
\label{sec:intro}
Over the past decades, advancements in radar technology have led to significant improvements in
resolution. Consequently, targets are no longer depicted as single points, but can be shown as a
two-dimensional grid. However, due to the high rotation rate of the radars and the vast surveillance
area, processing all targets simultaneously is computationally demanding. To address this challenge,
long-range radar systems often reduce the dimensionality of the detection grid for each target. The
one-dimensional data obtained as a result of this process are called \emph{High-Resolution Range
Profiles} (HRRP).

In defense-related research, collecting large and diverse measured datasets is challenging. Unlike
image classification, where abundant public data exist, radar datasets remain scarce. To address
this, researchers often rely on simulations, though these typically require complex modeling or
scaled targets \cite{img_gen_cond_hrrp}. Since HRRP data can be derived from more conventional radar
echoes, simulating them is simpler than emulating the full acquisition process. The MSTAR dataset is
frequently used for this purpose \cite{hrrp_ddpm, one_shot_gen, recog_aware, constrastive_learning_ma}, but it
includes only ~500 samples across ten targets, limiting its utility for machine learning. As a
result, most HRRP datasets involve only a few targets \cite{one_shot_gen, hrrp_clf_spectrum,
2dHRRP}.


One object can give rise to multiple HRRPs, depending on the acquisition scenario. Many targets,
such as ships or aircraft, remain underrepresented in large-scale HRRP databases, both in quantity
and diversity. Generative models can help fill these gaps by synthesizing missing objects and
scenarios, with the ultimate goal of enriching datasets for recognition. However, large-scale
studies on measured HRRP remain scarce, and foundational behaviors are not yet well documented. We
therefore focus on characterizing global trends in a real, large-scale HRRP database, leaving
task-specific recognition gains for future work. Our contributions can be summarized as
follows:

\begin{itemize}
\item \textbf{Foundational set of conditions.} We identify a minimal yet effective set of
      conditions that guide ship HRRP generation: ship dimensions and the desired aspect angle. 
      These parameters encapsulate key geometric properties, enabling the synthesis of coarse-scale 
      structural features.
\item \textbf{Geometry-driven insight at scale.} We recall a geometry-driven behavior in \emph{surface-radar} 
      HRRP data \cite{mfn}. Using a \emph{large-scale} dataset of ship measurements and metadata, we show 
      that conditional generative models consistently capture and reproduce this geometric trend using our set 
      of conditions.  
\item \textbf{Generalization across gaps.} Building on prior work \cite{multi_aspect_gen,one_shot_gen}, 
      our conditional models jointly address (i) \emph{missing scenarios} for known ships and (ii) 
      \emph{unseen ships}, enabling credible HRRP generation across these gaps.
\end{itemize}

%% file: 3_related.tex
\section{High-Resolution Range Profile Background}
\label{subsec:HRRPdata}
A radar system receives echoes of its transmitted waveform, which, after undergoing standard
processing steps \cite{richards2005fundamentals}, can be represented on a two-dimensional polar grid
of amplitudes $\sigma$, referred to as the \textit{radar cross section} (RCS). The amplitude is
maximized when a target effectively reflects the incident signal back toward the radar. This echo is
parameterized by two coordinates: the range $r$, corresponding to the distance from the radar, and
the azimuth $\phi$, corresponding to the angular position. A \textit{range profile} is obtained by
integrating $\sigma$ tangentially over a detection cone spanning the angular interval $[\Phi,
\Phi + \Delta \Phi]$. The spacing between consecutive range bins $r_i$ defines the radar range
resolution, hereafter denoted by $\Delta r$. The polar grid comprises $s$ cells in total, including
those activated by the target. Although the azimuth $\phi$ also possesses a finite resolution,
this component is absorbed into the amplitude of the curve after transformation into a
high-resolution range profile. Formally, the HRRP at range bin $r_i$ is given by $\mathrm{HRRP}(r_i)
= \sum_{\phi_j \in [\Phi, \Phi+ \Delta \Phi]} \sigma(r_i, \phi_j)$.

\begin{wrapfigure}[11]{l}{4.5cm}
    \centering
    \vspace{-6pt}
    \includegraphics[width=1.\linewidth]{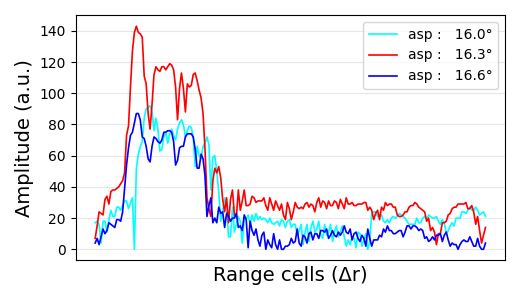}
    \caption{Three different HRRP measurements of the same ship at nearby aspect angles.}
    \label{fig:sensitivity}
\end{wrapfigure}
\vspace{4pt}

The structure of the HRRP is influenced by two angular parameters that characterize the acquisition
geometry: the \textit{aspect angle} and the \textit{depression angle}. The aspect angle quantifies
the relative orientation of the target with respect to the radar and is defined as the difference
between the target heading ($hdg$) and the radar azimuth ($\phi$), i.e., $\asp = hdg - \phi$.
The depression angle is the angle formed between the horizontal and the radar's \emph{line of
sight} (LOS). In \cite{multi_aspect_gen}, the authors show that a difference in aspect angle of one
degree can produce significant differences in the acquired signal. Our data reproduce the same 
phenomenon (Fig. \ref{fig:sensitivity}).


\begin{figure}[h]
    \centering
    \vspace{-5pt}
    \includegraphics[width=1\linewidth]{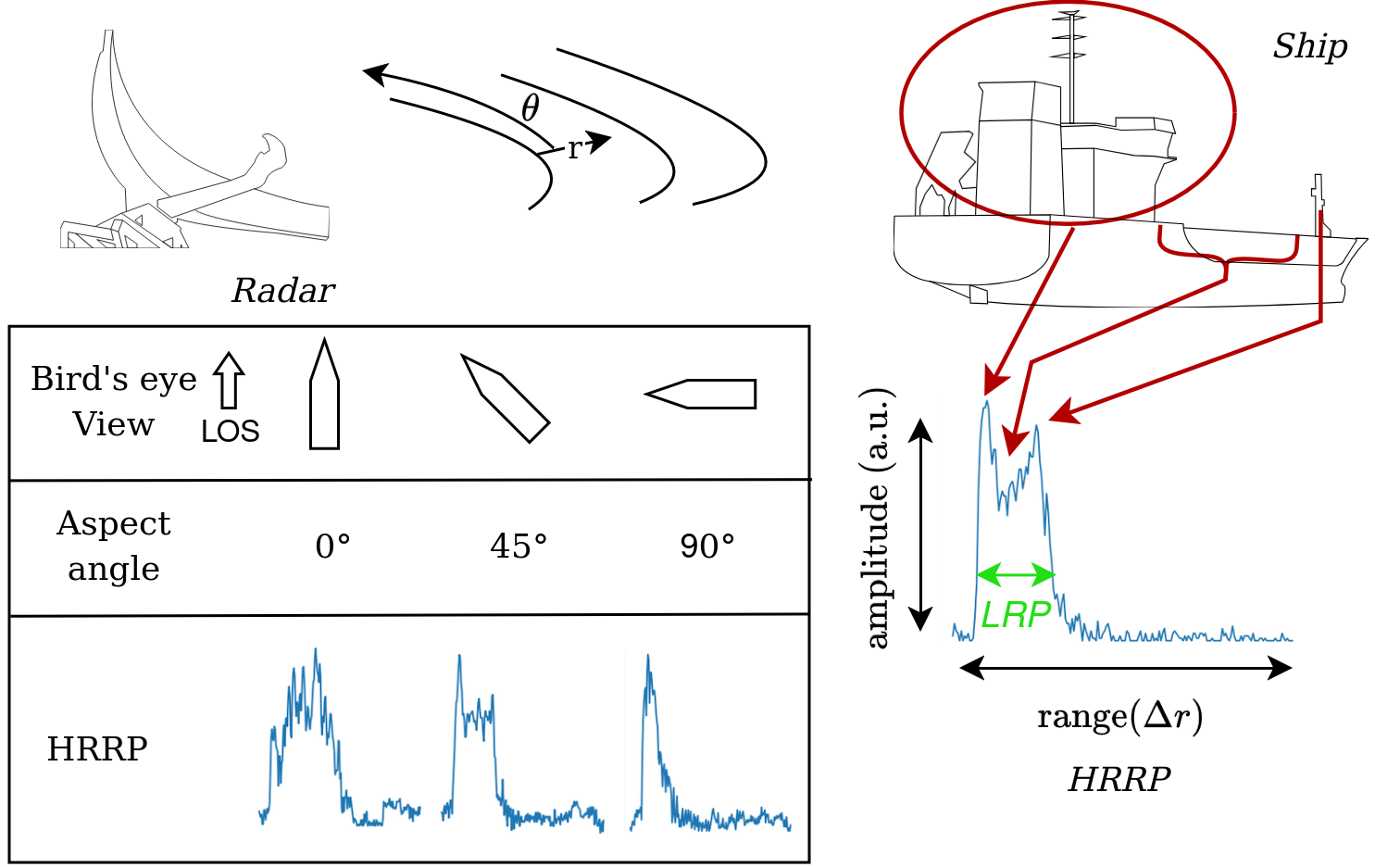}
    \caption{\textit{Illustration of HRRP structure and its dependence on aspect angle:} HRRP represents the amplitude of the combined echoes scattered by individual points within each range cell. Low-RCS parts of the ship cause drops in the HRRP, while high-RCS regions appear as peaks. The table highlights how aspect angle affects the coarse-scale structure of the HRRP by providing 3 data of the same ship at different aspect angles.}
    \label{fig:hrrp_def_sch}
    \vspace{-8pt}
\end{figure}

\vspace{5pt}


Numerous studies address HRRP-based target recognition, often using a single HRRP per target
\cite{hrrp_clf1,hrrp_clf2,hrrp_clf3}. However, given the high variability of HRRP signals
(Fig.~\ref{fig:sensitivity}) and their strong dependence on aspect angle
(Fig.~\ref{fig:hrrp_def_sch}), these approaches are hard to scale to large datasets. An alternative
is anomaly detection, which can focus on a specific target class and better generalize to unseen
targets \cite{bauw2020unsupervised}. The impact of aspect angle is critical: in
\cite{constrastive_learning_ma}, the authors leverage contrastive learning with data augmentation to
identify targets from HRRP captured at unseen views.

Classification performance ultimately depends on training data that reflect real-world diversity.
Due to the limited size of radar datasets, synthetic HRRP generation is often used to fill in
missing scenarios, most notably unseen aspect or depression angles
\cite{hrrp_ddpm, recog_aware, multi_aspect_gen}. However, these approaches remain limited, as they address
angular gaps but overlook more critical issues such as label imbalance, which significantly impacts
classification. A related effort in \cite{one_shot_gen} tackles the missing-target problem, but its
scope is hindered by a dataset of only three military vehicles, limiting both generalization and
practical relevance.

%% file: 4_method.tex
\section{Proposed Method}


\label{sec:format}
\subsection{Overview}

HRRP data are inherently challenging to interpret due to their sensitivity and the noise typically
present in radar measurements. These data correspond to projections of radar echoes onto the radar’s
LOS. A fundamental property that arises from this formulation is the variation of the ship’s
apparent length in the HRRP as a function of its aspect angle \cite{mfn}. This relationship can be expressed as
\begin{equation}
\label{eq:tlop}
TLOP(\length, \width, \asp) = \length|\cos(\asp)| + \width|\sin(\asp)|
\end{equation}
where Theoretical Length of Object Projection (TLOP) denotes the projection of the vessel on 
the radar’s LOS, while $\length$ and $\width$ correspond to the ship’s length and width, 
respectively. The parameter $\asp$ represents the vessel’s aspect angle at the time of HRRP 
acquisition.
\label{eq:lrp}
To recall the empirical TLOP in HRRP data as defined in \cite{mfn}, we first identify cells representing 
the ship's echo by smoothing the signal with a uniform filter, thresholding at 50\% of the maximum amplitude, 
and closing gaps shorter than 14 cells to keep the object contiguous. This yields the empirical Length on Range 
Profile (LRP) as the span between the first and last selected bins. The threshold and gap size are chosen 
to balance noise rejection. We empirically verified that this process yields a robust LRP estimate. The strong 
correlation between LRP and TLOP (Fig.~\ref{fig:theo_prac_lrp}) demonstrates the validity of this approach.

As a first step toward faithful HRRP synthesis, we aim to match
the TLOP distribution observed in real data; reproducing this
coarse geometric statistic is a prerequisite for any further realism.

\subsection{Conditioning Generative Models for HRRP}

We compare two generative paradigms: Denoising Diffusion Probabilistic Models (DDPM) \cite{ddpm} and Generative
Adversarial Networks (GAN) \cite{gan} for HRRP synthesis. Our objective
departs from conventional unconditional generation: rather than maximizing generic sample diversity,
we seek controlable generation tied to target attributes. To this end, we condition the model on ship length, width, and aspect angle
(collectively represented by the vector $c$), which enforces the global structure of the generated
HRRP. In both architectures, conditioning is applied by adding learned embeddings between every 
downsampling and upsampling stage. 


\subsubsection{Denoising Diffusion Probabilistic Models (DDPMs)}
Building on the classifier-free conditioning approach \cite{classifier_free}, our training process
employs the following loss function
\begin{equation}
    \mathcal{L}_{cddpm} = \hspace{-6pt} \mathop{\mathbb{E}}_{t, x_0, \epsilon, c} \left[ \left\| \epsilon - \epsilon_\theta \left(\sqrt{\bar{\alpha}_t} x_0 + \sqrt{1 - \bar{\alpha}_t} \epsilon, t, c \right) \right\|^2 \right]
\end{equation}

\subsubsection{Conditioning GANs}
We use a Wasserstein GAN (WGAN) \cite{arjovsky2017wasserstein} as a second generative model
architecture. In the following equations, $q$ denotes the data distribution and $p$ the latent distribution. 
We train the WGAN by computing alternatively the generator loss,
\begin{equation}
    \mathcal{L}_{advG} = - \mathop{\mathbb{E}}\limits_{z \sim p, c} [\discriminator(\generator(z, c))]
\end{equation}
and the discriminator loss
\begin{equation}
\mathcal{L}_{advD} = \mathop{\mathbb{E}}\limits_{z \sim p, x \sim q, c} [\discriminator(\generator(z,c)) - \discriminator(x)].  
\end{equation}
We use a Wasserstein GAN loss formulation. We enforce the critic’s 1-Lipschitz constraint using weight clipping by clamping each parameter
to the range $[-0.05, 0.05]$ after every discriminator update. Also, we augment the adversarial objective 
with a mean-squared-error (MSE) reconstruction term to promote signal-level fidelity between the real 
data \( x \) and the generated samples \( \generator(z, c)\). The MSE loss is defined as
\begin{equation}
    \mathcal{L}_{mse} = \mathop{\mathbb{E}}_{z \sim p, \, x \sim q, \, c} \left[\left\| x - \generator(z, c) \right\|^{2} \right].
\end{equation}


To balance the contributions of each loss term, we introduce a weighting factor \(\lambda\).
Since the data are normalized, MSE loss is typically smaller in magnitude compared to
the adversarial loss. Therefore, we set \(\lambda = 50\) in all experiments. The discriminator is
trained with \(\mathcal{L}_{advD}\), while the generator minimizes
\begin{equation}
    \mathcal{L}_{totG} = \mathcal{L}_{advG} + \lambda \mathcal{L}_{mse}.
\end{equation}


%% file: 5_results.tex
\section{Experimental Setup and Findings}

\subsection{Dataset}

\label{subsection:dataset}
We leverage a large database of ship HRRP data acquired by a coastal radar system over several months.
This dataset contains over 900k HRRP samples from more than 700 unique ships, covering a wide 
range of acquisition scenarios typical of coastal surveillance. Each HRRP is associated with 
metadata including the ship's Maritime Mobile Service Identity (MMSI), dimensions (length and width) 
and the aspect angle at acquisition, defined as the angle between the ship's heading and the radar's 
line of sight.

We collect our raw data using a ground-based radar located slightly above sea level. Given its
elevation (around 100m) and the typically much larger range to targets, we consider the impact of
the depression angle on HRRP data to be negligible. We therefore focus on the effect of the aspect 
angle, as it significantly influences the radar cross-section and the resulting HRRP.

\begin{wrapfigure}[9]{l}{5cm}
    \includegraphics[width=1.\linewidth]{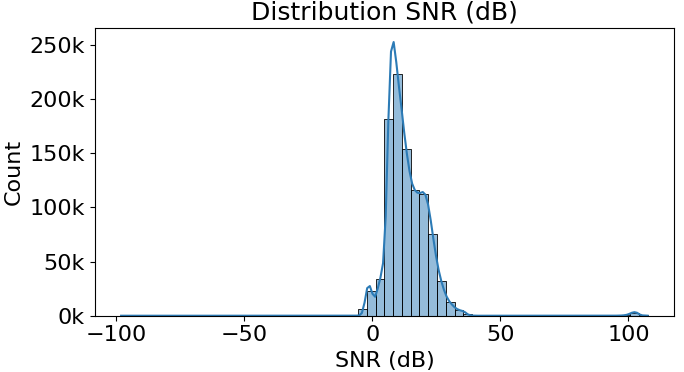}
    \caption{Dataset SNR distribution.}
    \label{fig:snr_distribution}
\end{wrapfigure}

\/*
\begin{figure}[h]
    \centering
    \includegraphics[width=1\linewidth]{Models_archi.png}
    \caption{Detailed architecture of proposed models. The same conditioning method is used in both
models. The variable s represents the size of HRRP data.}
    \label{fig:archi}
\end{figure}
*/

Using the LRP calculation from Sec. ~\eqref{eq:lrp}, we estimate for each HRRP data the noise part and 
the signal part of the data. We then compute the SNR as the ratio between the average power of the
signal part and the average power of the noise part. Figure \ref{fig:snr_distribution} shows the SNR distribution in our dataset.
We observe that most data have an SNR between 10dB and 30dB, with an average of 13dB, indicating a
reasonable signal quality for our generative tasks.

\begin{wrapfigure}[18]{r}{4cm}
    \centering
    \vspace{-8pt}
    \includegraphics[width=1\linewidth]{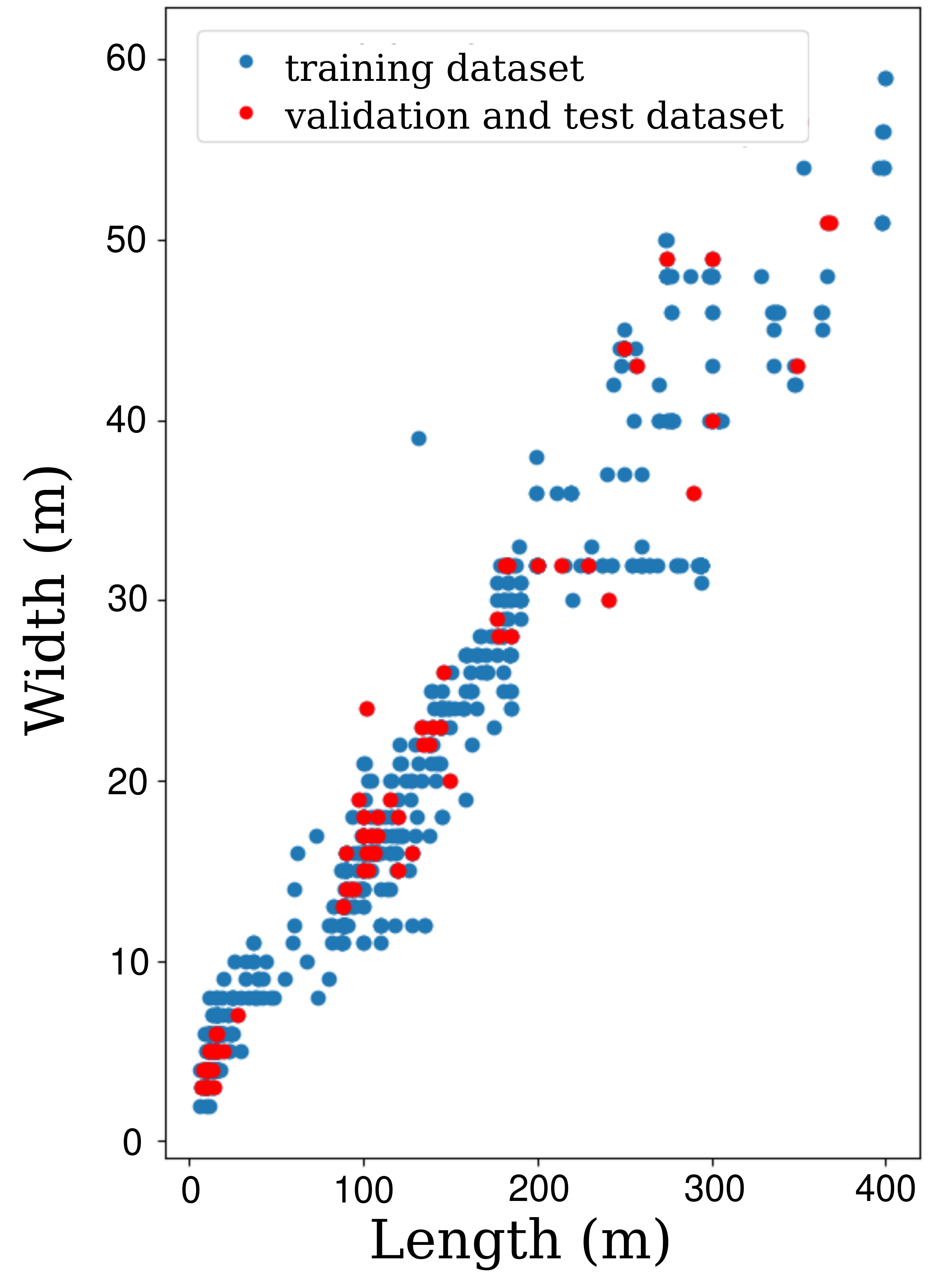}
    \caption{Dataset ship dimensions. \textbf{Blue:} train, \textbf{Red:} val/test.}
    \label{fig:vt_range}
\end{wrapfigure}

We use the MMSI (a unique identifier for each ship) to select specific ships for the test dataset to ensure no overlap with the training data. 
In all experiments, the training dataset represents $90\%$ of the entire dataset, and the validation and 
test datasets represent half of the remaining $10\%$ of MMSI (still containing 100k data). We randomly 
select MMSI to fill the test/validation datasets and select the same way half of this set to fill in the 
validation dataset and the other half for the test dataset. Figure \ref{fig:vt_range} shows the distribution 
of ship lengths and widths in the train and val/test datasets.

We adopt lightweight architectures to limit overfitting on low-dimensional HRRP data. The proposed GAN 
has 697k parameters (including 146k for the discriminator), and the DDPM has 730k. Both models are 
built upon 1D convolutional ResBlocks \cite{resnet}. Aspect angles are encoded with sinusoidal embeddings, concatenated with the length and width, mapped
through linear layers, and injected into the network by addition to the feature maps between successive 
downsampling and upsampling stages. We use a cosine noise schedule with 800 diffusion timesteps throughout 
training, which helps balance sample quality and computational efficiency.

\subsection{Evaluation Metrics}

Previous works on HRRP generation \cite{hrrp_ddpm,one_shot_gen,multi_aspect_gen} primarily
assess the benefit of generated data through classification accuracy improvements. In contrast, our goal  
conditions do not carry class information, classification would therefore not be meaningful. Thus, we
evaluate generation fidelity using the metrics from \cite{mfn}, also enabling cross-paper comparison.

Due to the high sensitivity of HRRP to aspect angle (Fig.~\ref{fig:sensitivity}), pointwise
comparisons with a single reference are unreliable. Instead, for each test sample, we define a local
set of real HRRP within a small angular window \([\asp-\Delta,\ \asp+\Delta]\), compute metrics
between each generated data and all elements in this set, retain the best match, and average these best scores over the test
set. A generation is considered successful if it closely matches any real trace in its
neighborhood. In practice, we set $\Delta=2^\circ$ to define a narrow neighborhood. We generate 
a single sample per condition and select the closest real HRRP within a ±2° neighborhood; thus, 
the “best match” is taken over real neighbors, and full distribution
matching is left for future work.

We adopt the $\MSE_f$ and $\cos_f$ metrics from \cite{mfn}, and additionally compute PSNR
restricted to activated cells in order to align with other metrics. The $\cos_f$ metric quantifies the fidelity of scatter patterns within the signal, while
$\MSE_f$ and PSNR provide global measures of similarity between the generated data and the
ground-truth neighborhood. As the $\MSE_f$ is the most general metric, 
we use it to define the closest match in the neighborhood for PSNR and $\cos_f$ calculations.


\subsection{Results}

\begin{figure*}[t]
    \centering
    \includegraphics[width=0.8\linewidth]{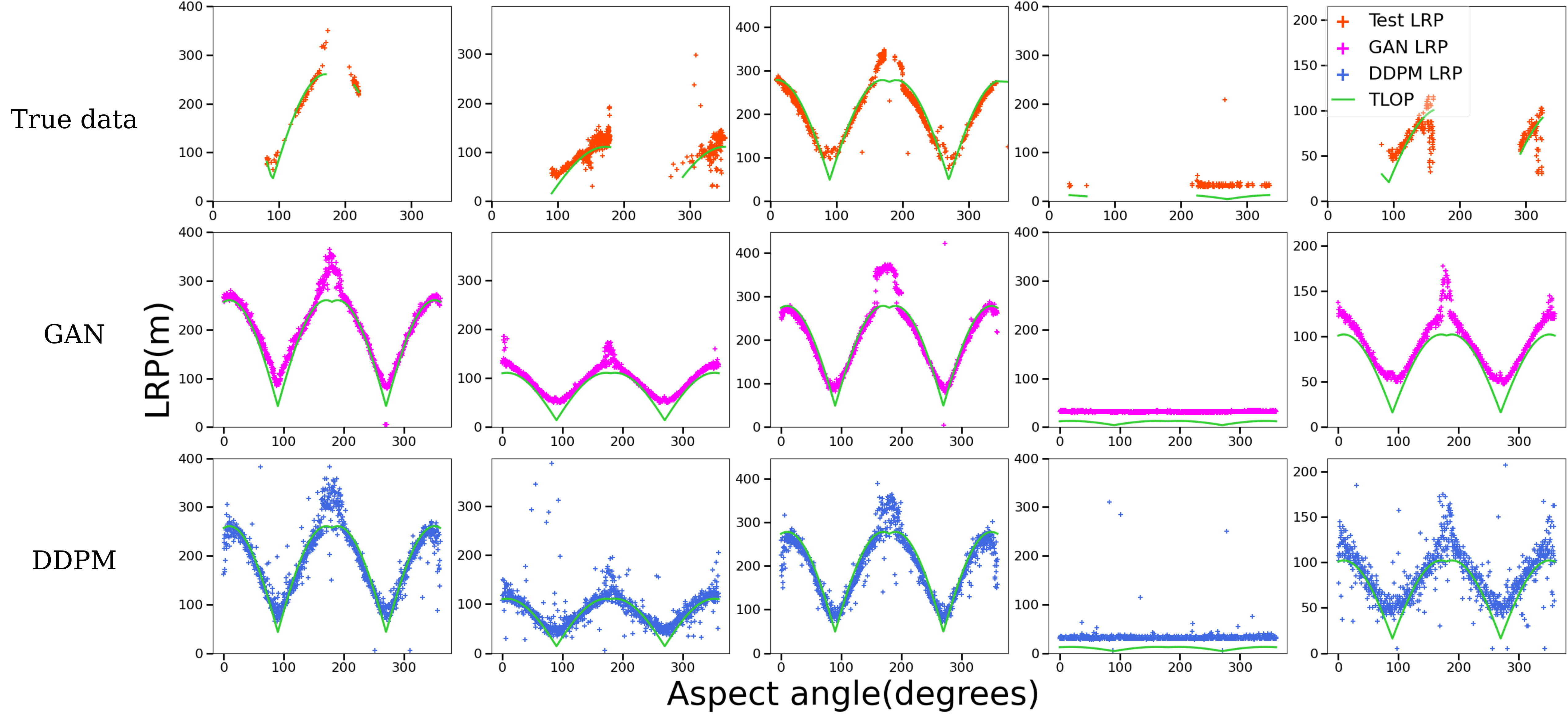}
\caption{LRP profiles from test data and samples generated with the proposed models. 
\textbf{Top:} test LRPs; gaps in aspect angles reflect missing measurements. 
\textbf{Middle:} GAN-generated LRPs conditioned on aspect angle, which follow the main trend but do
not reproduce outliers. \textbf{Bottom:} LRPs measured on DDPM-generated HRRP conditioned on aspect
angle, capturing both the trend and the outliers. \textbf{Green:} theoretical ship projection
(TLOP, Eq.~\eqref{eq:tlop}). The different columns correspond to different ships of various dimensions. 
Overall, LRPs from generated and test data follow the TLOP trend.}
    \label{fig:theo_prac_lrp}
\end{figure*}

As shown in Fig.~\ref{fig:theo_prac_lrp}, for both architectures, the $LRP$ correlates strongly with 
the theoretical projection ($TLOP$). GAN-generated profiles follow the $TLOP$ trend across aspect 
angles but miss the dataset outliers. DDPM-generated HRRP display a realistic trade-off: they 
capture both the outliers and the main trend. The test data departs from the $TLOP$ for 
aspect angles of 180\textdegree, likely due to the influence of the bridge of the ship, not 
accounted for in the simple projection model. This effect is well captured by the generative models. 
For very small ships, the $TLOP$ model is less accurate, certainly due to the relatively large resolution
of the radar system compared to the ship size. This effect is also well captured by the generative models.
\/*
\begin{figure}[h]
    \centering
    \includegraphics[width=0.8\linewidth]{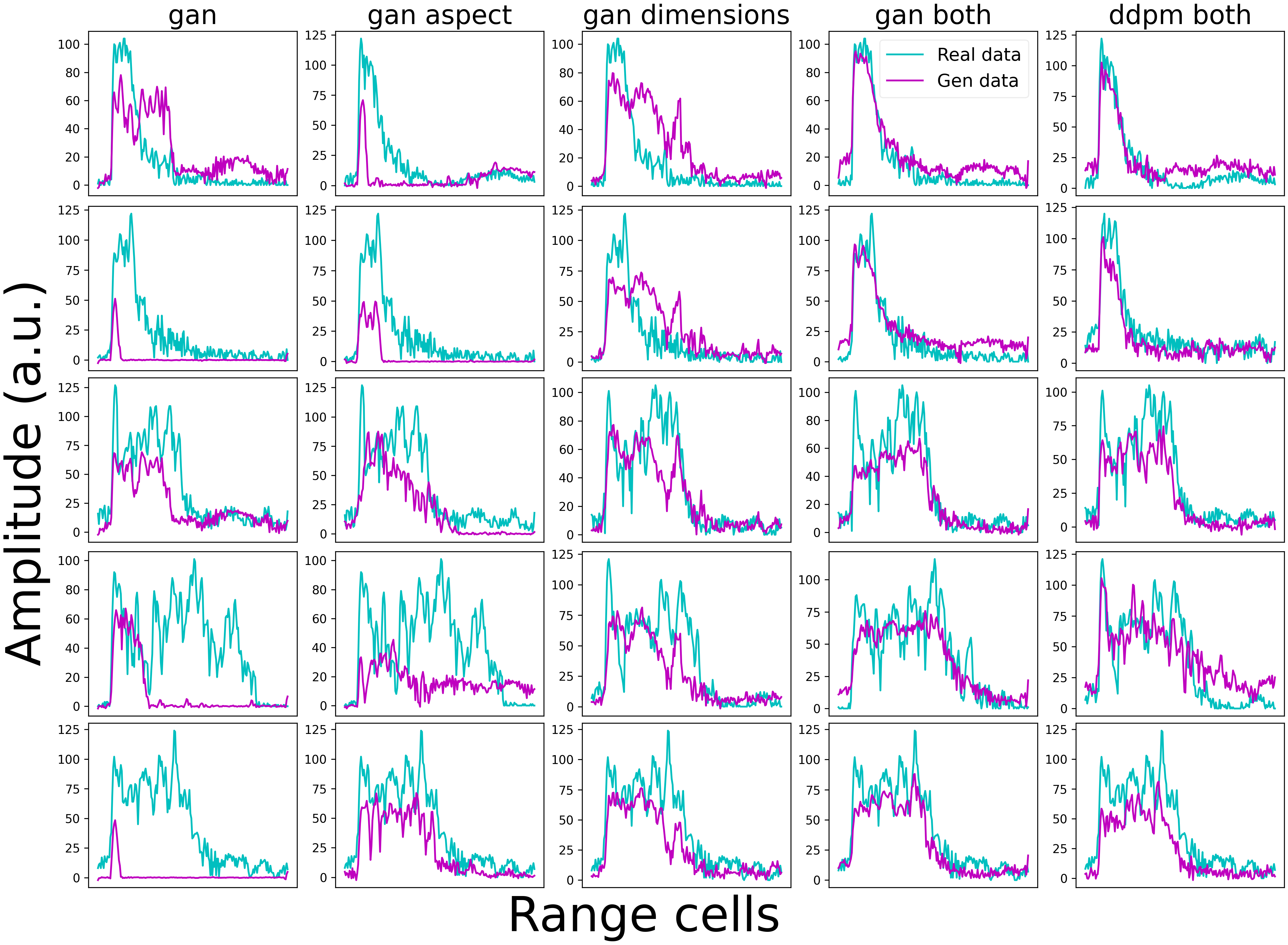}
    \caption{Experiment $(A)$: Comparative figure of the generated HRRP (\textbf{cyan}) and the
    closest RP data of the same ship in the dataset within the 5° aspect angle tolerance introduced in
    \ref{cinqdeg} (\textbf{magenta}).}
\end{figure}
*/
\begin{table}[ht]
\caption{Generation metrics for different models and conditioning types. The best scores for each
model are in \textbf{bold}.}
\label{tab:results}
\centering
\begin{tabular}{l l c c c}
\toprule
Model & Conditioning & PSNR $\uparrow$ & MSE$_f \downarrow$ & cos$_f \uparrow$ \\
\midrule
\multirow{4}{*}{GAN}
  & none & 15.7 & 11.6 & 0.33 \\
  & aspect angle & 17.4 & 9.2 & 0.42 \\
  & dimensions & 23.5 & 3.2 & 0.66 \\
  & aspect \& dimensions & \textbf{27.0} & \textbf{0.95} & \textbf{0.87} \\
\midrule
\multirow{4}{*}{DDPM}
  & none & 15.8 & 13.0 & 0.38 \\
  & aspect angle & 16.3 & 12.5 & 0.43 \\
  & dimensions & 22.0 & 4.31 & 0.62 \\
  & aspect \& dimensions & \textbf{24.7} & \textbf{1.51} & \textbf{0.81} \\
\bottomrule

\vspace{-15pt}
\end{tabular}
\end{table}
Table \ref{tab:results} presents the generation metrics obtained in our experiments. We observe that 
conditioning with the dimensions, providing a strong prior on the HRRP structure, has the most
significant impact on generation quality. However, adding aspect angle conditioning further improves
performance, demonstrating that both conditions are complementary. Both models show similar trends, 
demonstrating this behavior is not method-dependent. Overall, the GAN achieves higher scores, 
likely due to the smoothing effect of the MSE loss, as discussed below.

\begin{figure}[t]
    \centering
    \includegraphics[width=1.\linewidth]{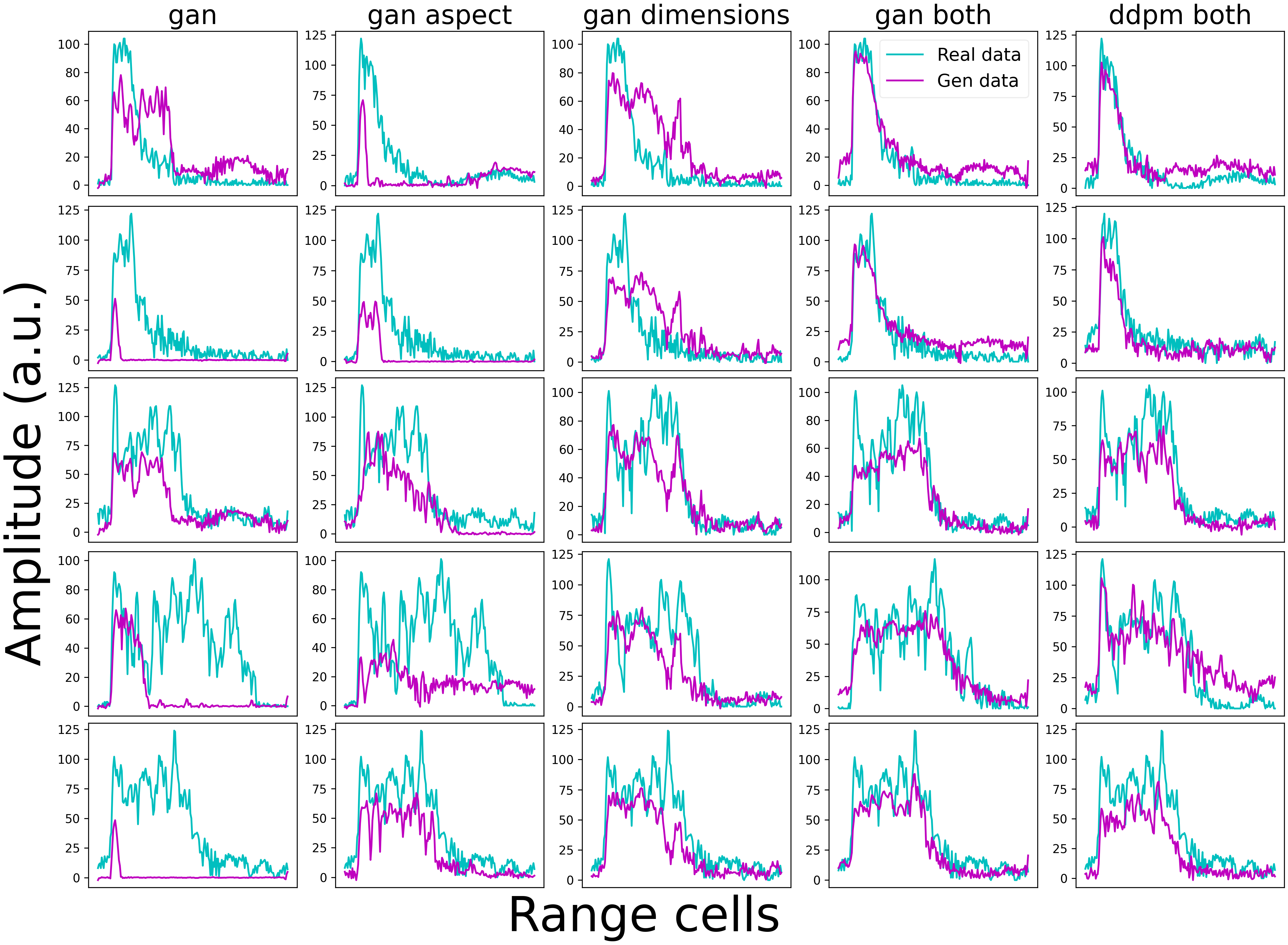}
    \caption{Comparative figure of the generated HRRP (\textbf{cyan}) and the closest RP data of 
    the same ship in the dataset within the 2° aspect angle tolerance (\textbf{magenta}). The columns 
    correspond to different models and conditioning, and the rows to different ships. We observe that
    before conditioning with both the aspect angle and dimensions, the overlap between generated and
    real data is lower than using both.}
    \label{fig:hrrp_comparison}
\end{figure}

We provide a comparative figure of the generated data and measured data (Fig. \ref{fig:hrrp_comparison}). 
The GAN model with full conditioning provides a similar match to the DDPM, but its signal is smoother. 
This phenomenon is likely due to the MSE loss used in the GAN training, which encourages averaging over 
possible outputs. On the contrary, the DDPM model generates signals with more variability, closer to 
real data but the smoothing phenomenon helps the GAN achieving higher metrics scores. Overall, both models 
using all conditions generate HRRP data with similar structures to real data of the same ship yet missing 
some fine details because of the coarse conditioning.

\subsection{Discussion}

We do not target ship identification in this work. The models are conditioned only on coarse
geometry (length, width, aspect angle) and deliberately avoid identity cues, so naive data
augmentation for classification could introduce label noise. Our aim here is to characterize and
reproduce global, geometry-driven structure in HRRP at scale, providing a foundation for future
recognition-aware extensions using richer conditioning.

This study highlights that, despite the sensitivity of HRRP data to acquisition conditions,
consistent patterns can still be learned and exploited in large-scale HRRP datasets. 
Our generative models leverage shared features among ships with similar dimensions, but 
incorporating richer conditioning could further improve fidelity. To support future work, 
we release a subset of modified data and generated samples for reproducibility, available 
at: \href{https://github.com/EdwynBrient/HRRPGen-geometry}{https://github.com/EdwynBrient/HRRPGen-geometry}.



%% file: 6_conclusion.tex
\section{Conclusion}
We present the first study on HRRP generation using a large-scale
dataset capturing the diversity of coastal radar surveillance
scenarios. Our analysis highlights the mutual dependence between
the conditioning variables, namely the target dimensions and aspect angle. 
The generated signatures reproduce the TLOP geometric property observed 
in real HRRP data. Finally, our conditional models achieve promising 
performance while relying on conditions that do not uniquely characterize 
the target, underscoring the fundamental role of these variables in HRRP synthesis.